\documentclass[letterpaper, 10 pt, journal, twoside]{IEEEtran}
\usepackage[pdftex]{graphicx} 
\usepackage{gensymb}
\usepackage{amsmath}
\usepackage{hyperref}
\IEEEoverridecommandlockouts  
\usepackage{booktabs}
\usepackage{xcolor}
\usepackage[normalem]{ulem}
\usepackage{tikz}
\usepackage{xfrac}
\usepackage{float}
\usepackage{placeins}
\usepackage[justification=centering]{caption}
\usepackage[pscoord]{eso-pic}
\begin{document}

\title{Towards Sustainable Development: A Novel Integrated Machine Learning Model for Holistic Environmental Health Monitoring}
\author{
\begin{tabular}{c} Anirudh Mazumder \\ Texas Academy of Mathematics \& Science \\ Denton, United States of America \\ anirudhmazumder26@gmail.com \end{tabular} \and
\hspace{50pt} 
\begin{tabular}{c} Sarthak R. Engala \\ Texas Academy of Mathematics \& Science \\ Denton, United States of America \\ sre2048@gmail.com \end{tabular} \and
\hspace{50pt}
\begin{tabular}{c} Aditya Nallapuraju \\ Texas Academy of Mathematics \& Science \\ Denton, United States of America \\ aditya.nallaparaju@gmail.com \end{tabular}
}
 
\maketitle
\begin{abstract}
Urbanization enables economic growth but also harms the environment through degradation. Traditional methods of detecting environmental issues have proven inefficient. Machine learning has emerged as a promising tool for tracking environmental deterioration by identifying key predictive features. Recent research focused on developing a predictive model using pollutant levels and particulate matter as indicators of environmental state in order to outline challenges. Machine learning was employed to identify patterns linking areas with worse conditions. This research aims to assist governments in identifying intervention points, improving planning and conservation efforts, and ultimately contributing to sustainable development.
\end{abstract}

\begin{IEEEkeywords}
Machine learning, environmental sustainability, air quality, water quality
\end{IEEEkeywords}

\section{Introduction}
Urbanization has become a global phenomenon in the 21st century, marking a new era of global modernity \cite{Kookana2020}. The migration of people to urban centers has brought about significant economic opportunities, increased social mobility, and improved infrastructure in the form of transportation, sanitation, and technology networks \cite{Pradhan2021}. However, rapid and uncontrolled urbanization has also resulted in intensive environmental exploitation and degradation, posing a complex sustainability challenge for municipalities \cite{Yang2022}. Traditional methods of detecting environmental problems rely heavily on direct observation, manual data collection, and periodic reporting, which are often inefficient, costly, and provide only fragmentary insights. Consequently, separate indices have been created for different environmental quality metrics, such as air and water quality \cite{Wang2022}, to address these issues; however, there have yet to be concerted efforts to integrate these indices to provide a comprehensive, holistic view of overall environmental health.

Machine learning techniques hold promise for addressing complex classification problems involving multidimensional, evolving data streams \cite{Pugliese2021}. Using traditional statistical methods, machine learning models can identify intricate patterns that are not easily detectable. They also can adapt to new data over time continuously. These characteristics make machine learning well-suited for developing scalable, flexible classification systems. Given environmental health data's complex, multifaceted nature, machine learning presents a promising avenue for developing an integrated environmental classification model to promote effective sustainability planning.

Therefore, it was hypothesized that a machine learning model could be developed to classify environments into different labels that holistically define overall environmental health, integrating across the distinct indices for water, air, soil, biodiversity, and other facets, which would help address the key limitations of existing environmental monitoring methods, relying on fragmented indices and incomplete data sources. Current methods also often require extensive manual data collection and analysis, which is inefficient and costly. A machine learning model could be trained on large, multidimensional environmental datasets to identify intricate patterns and relationships between different aspects of environmental health. Such a model could provide a more holistic, automated assessment of current conditions and emerging threats. Additionally, machine learning techniques can continuously adapt to new data over time, enabling the model to detect novel environmental issues that static statistical methods may miss. Overall, the strengths of machine learning in handling complex, evolving classification problems make it a promising approach to pursue developing an integrated, efficient, and proactive environmental health monitoring system.
\section{Methodology}
\subsection{Materials}
Python was the primary programming language utilized for developing and testing the machine learning algorithm throughout the research. Indian Air Quality and Indian Water Quality datasets were the primary data sources.
\subsection{Pipeline}
\subsubsection{Algorithm}
\begin{figure}[!htb]
	\centering
	\includegraphics[width=\columnwidth]{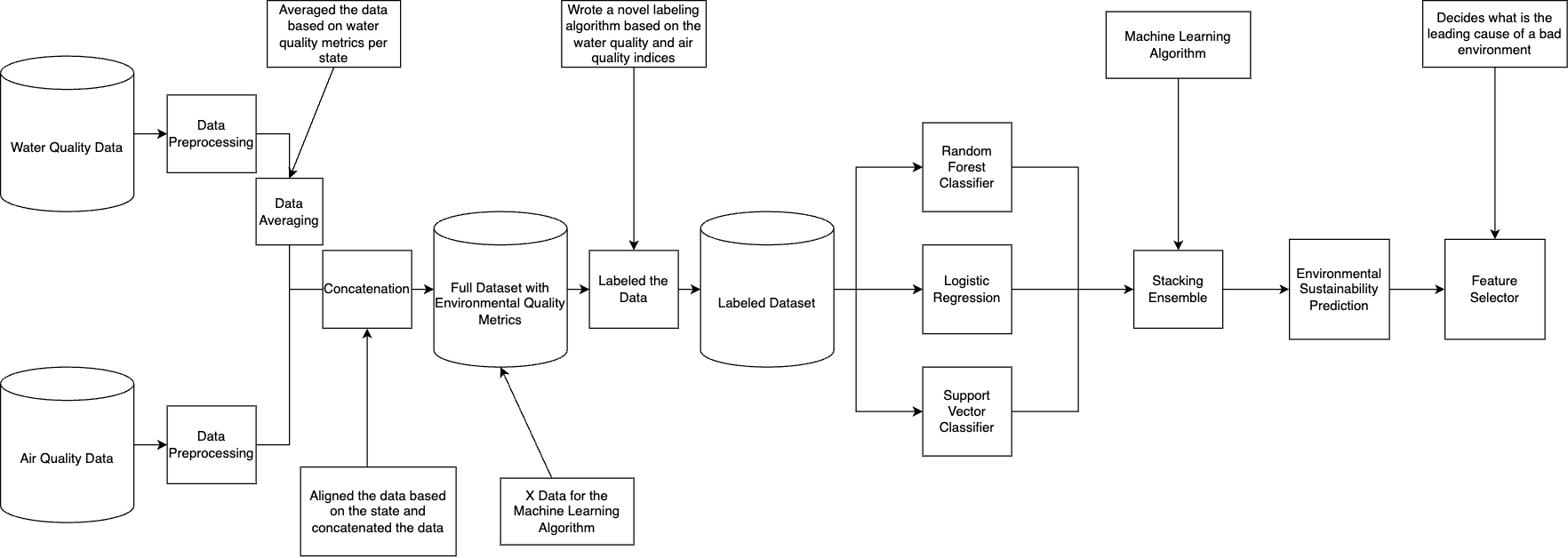}
	\caption{Flow chart showing the novel pipeline that was built}
	\label{fig: Figure 1}
\end{figure}

The machine lear˜ning pipeline was split up into multiple different steps, to train a machine learning model which is able to generate high accuracy and precise outputs. The layout of the different steps can be seen in Figure \ref{fig: Figure 1}
\FloatBarrier
\FloatBarrier
\subsubsection{Data Preprocessing}
The algorithm's initial stage involved preprocessing the data in the datasets. For the Indian Air Quality dataset, the primary metrics were chemical concentrations and particles present in the air. These metrics were utilized to calculate the air quality index for each data point in the dataset. The same process was repeated for the Indian Water Quality dataset to enable the classification of air and water quality.
\subsubsection{Concatenation}
To ensure the alignment of the data, a similarity between the datasets was identified, which was found in the location of the metrics within the datasets. Since the alignment was based on the state, one of the datasets needed to be averaged for each metric by state. In this case, the water quality dataset was chosen for averaging as it contained more data for each state, resulting in a more representative and generalizable value for the dataset compared to the air quality dataset. After averaging the water quality dataset by state, it was aligned with the air quality dataset. This alignment process was crucial to facilitate the concatenation of the two datasets, ensuring that random state datasets were not simply combined. Once the data was aligned, the datasets were concatenated to create a unified dataset suitable for our specific task.
\subsubsection{Labeling}
In order to create a machine learning model capable of accurately assessing the environmental challenges faced by cities, relevant data needed to be labeled. Without labels, the model would have had to rely solely on unsupervised learning techniques, which allow models to find patterns in data without guidance. However, supervised learning models are more accurate for specialized classification tasks \cite{JMLR:v24:22-0019}. To enable supervised learning, data about cities worldwide was manually reviewed, and each city was assigned a label describing its environmental conditions. Six labels were defined, ranging from very negative environmental conditions (severe) to very positive ones (good). The labeling process involved examining factors related to water quality, including contaminant levels, pollution sources, and treatment methods, as well as air quality data covering pollution levels, emissions sources, and respiratory illness rates. By comparing the labeled dataset, the model could learn which environmental characteristics matched each descriptive label, allowing it to classify new, unlabeled city data based on patterns from the labeled examples. The labels enabled supervised learning that produced a significantly more accurate classification model than unsupervised methods could have.
\subsubsection{Machine Learning}
To effectively learn from the data, the dataset was split into two sets: 80\% for training and 20\% for testing. A stacking ensemble model was employed, consisting of three machine learning models: Random Forest Classifier (RFC), Support Vector Classifier (SVC), and Logistic Regression. The RFC was chosen for its ability to utilize multiple decision trees, specified to be ten predictors, to predict the best classifier based on a list of metrics about the environment. The SVC was selected for its ability to separate all data points into their respective classes based on the feature space, allowing the algorithm to identify outliers and prevent them from influencing the decision made by the algorithm. Lastly, the logistic regression model was chosen for its ability to understand how many independent variables can influence a specific dependent variable, making it an accurate predictor of the label to which the data is classified.

The unique abilities of these different models were combined to maximize the predictor's accuracy. Ensemble models, which combine multiple models to create a more robust prediction model, were employed. Since the base models were all different machine learning algorithms, a stacking-based ensemble was performed. This approach finds the optimal combination of multiple other machine learning models and can utilize multiple machine learning algorithms. A stacking algorithm was used instead of other ensembling methods because it can utilize all the different machine learning algorithms while other models cannot. Additionally, it is based on a hierarchical structure rather than a flat equal-level model, which is essential because it allows the algorithm to use the strengths of all the base models. Most importantly, stacking algorithms can improve the accuracy of the overall machine-learning algorithm.
\subsubsection{Feature Selector}
To perform feature selection, the labels were converted into numerical representations. Specifically, the values 1 were assigned for "good," 2 for "satisfactory," 3 for "fair," 4 for "poor," 5 for "bad," and 6 for "severe." This numerical representation allowed for the computation of the correlation between the highest values and the features contributing to this correlation \cite{Sedgwick12}.
\subsection{Labeling Algorithm}
The dataset used in the study had two unique features: the water quality index and the air quality index. A labeling classifier was built to develop a specific environmental quality classifier based on these features. The air quality index was classified into six categories: good, satisfactory, moderate, poor, very poor, and severe. In contrast, the water quality index was classified into five categories: excellent, good, satisfactory, poor, and severe.

To create a holistic label that describes the environment, an approximate average between the feature values from the two labels was calculated. The labeler was classified using a specific structure, which involved considering the following scenarios:
\begin{itemize}
\item If air and water quality are the same (e.g., both 'Good'), the overall classification is set to the same value (i.e., 'Good' in this case).
\item If air quality is 'Good' and water quality is 'Excellent,' the overall classification is set to 'Excellent.'
\item If air quality is 'Good' and water quality is 'Good,' the overall classification is set to 'Good.'
\item If air quality is 'Good' and water quality is 'Satisfactory,' the overall classification is set to 'Good.'
\item If air quality is 'Good' and water quality is 'Poor,' the overall classification is set to 'Fair.'
\item If air quality is 'Good' and water quality is 'Severe,' the overall classification is set to 'Bad.'
\end{itemize}
The model incorporated both air quality and water quality labels, but with a higher weight placed on air quality. This weighting scheme was used because the air quality labels provided more granular scoring compared to the broader water quality labels. The air quality data enabled finer distinctions between different pollution levels. In contrast, the water labels categorized water quality in broader terms. Since the air quality data offered greater complexity and ability to discern subtle differences, it received a higher weightage in the model. This helped account for its higher granularity while still allowing the model to take advantage of its informative scoring. At the same time, the water quality labels provided useful complementary signals, just weighted lower than air quality. By giving air quality higher influence, the model could benefit from its granular detail while seamlessly integrating the water quality data as well. The weighted, multi-faceted labeling enabled robust predictions that took advantage of the strengths of both data sources.
\section{Results}
\subsection{Accuracy}
\begin{figure}[!htb]
	\centering
	\includegraphics[width=\columnwidth]{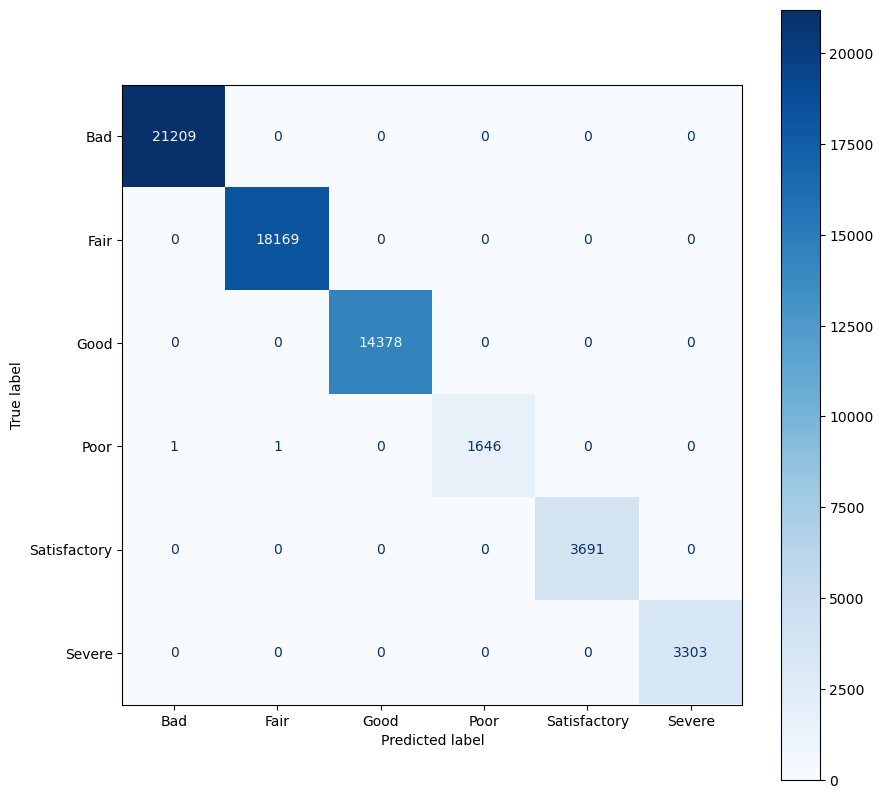}
	\caption{Confusion matrix showing the predicted vs the actual values}
	\label{fig: Figure 2}
\end{figure}

To rigorously evaluate the performance of the proposed machine learning model, a standard train-test split technique was utilized. The original dataset was divided such that 80\% of the data was allocated to a training set for model optimization, while the remaining 20\% was held out as a test set. This testing data was not used in model training and provided an unbiased estimate of generalizability and accuracy. The trained model was applied to the test set, and predictions were generated for all test points. These predictions were compared to the known ground truth labels, allowing the calculation of the test accuracy. The model achieved an overall accuracy of 0.99\% on the unseen test data. This high accuracy indicates excellent generalization ability and suggests that the patterns learned during training generalize very well to new, unseen data. To gain further insight, a confusion matrix was generated to break down the prediction performance for each class. As seen in the total confusion matrix in Figure \ref{fig: Figure 2}, the exceptionally high diagonal and low off-diagonal values confirm the high accuracy across all classes. Out-of-sample testing helps prevent overfitting, and the decisive test performance demonstrates that the model has learned robust representations that go beyond just fitting the training data. This rigorous train-test evaluation provides confidence that the model will maintain high accuracy when deployed in practice. It also highlights the importance of proper evaluation techniques to ensure that the model's performance is not overestimated.
\FloatBarrier
\FloatBarrier
\subsection{Feature Selector}
\begin{figure}[!htb]
	\centering
	\includegraphics[width=0.9\columnwidth]{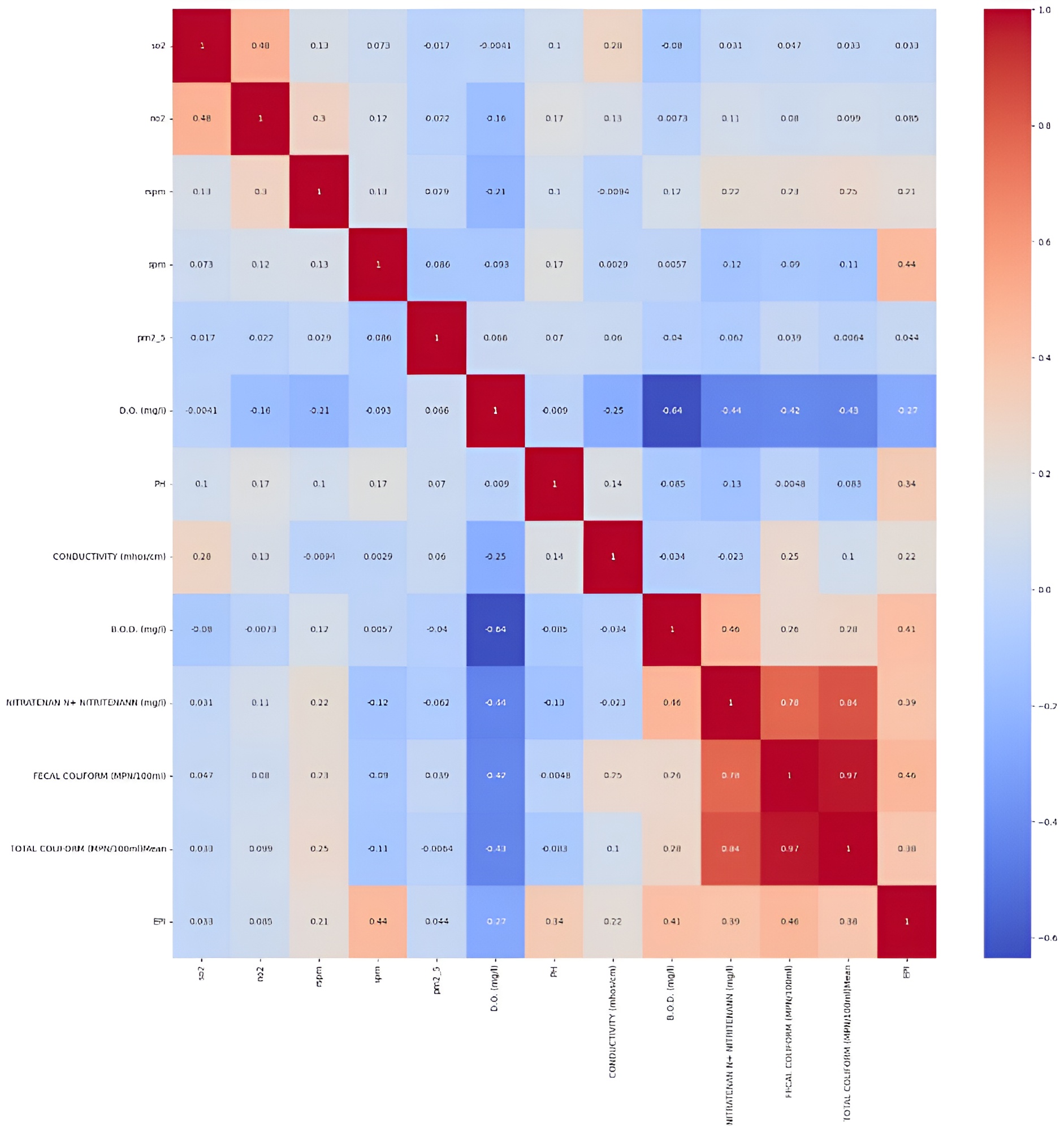}
	\caption{Feature selection matrix showing correlations}
	\label{fig: Figure 3}
\end{figure}

In this research, a feature selection was performed using a correlation-based approach to determine which water and air quality parameters were most predictive of an unfavorable environment. The Pearson correlation coefficient of each feature with the target variable of bad water quality was computed, and features were ranked based on the absolute value of the correlation which can be seen in Figure \ref{fig: Figure 3}. The feature selection results showed that fecal coliform levels had the highest positive correlation with bad water quality, indicating that higher fecal coliform counts are associated with a deteriorating environment. Total suspended solids (spm) and biochemical oxygen demand (BOD) were the most correlated parameters, followed by nitrate, total coliform, pH, and conductivity. Notably, dissolved oxygen (DO) exhibited a negative correlation, implying that lower DO relates to worsening conditions.

Feature selection is a critical modeling step that can significantly improve the performance and interpretability of environmental quality assessment. By systematically ranking the predictive capacity of each feature, one can focus on just those variables that have the strongest relationship with environmental health. This data-driven approach helps weed out all the non-informative "noise" features that offer little signal.
Identifying the critical predictive indicators provides vital insights that can guide targeted remediation efforts to improve environmental conditions. For example, knowing fecal coliform is strongly associated with degraded water environments suggests wastewater treatment interventions could have a high impact. Reducing organic pollution and nitrate levels also appears promising based on the strength of their correlations. Isolating the prime factors driving environmental deterioration allows policymakers and sustainability professionals to allocate resources more effectively to initiatives and technologies capable of addressing the most influential issues. The power of feature selection is not just improving predictive modeling but revealing actionable opportunities for meaningful interventions to restore the ecosystem and community wellbeing on a broader scale.
\FloatBarrier
\FloatBarrier
\section{Discussion}
\subsection{Conclusion}
As the world continues to urbanize at an unprecedented rate, the environment is deteriorating under the immense strains that cities place on natural resources, with no way to quantify or identify which areas are most at risk accurately. Expanding populations and industrialization results in polluting air and waterways. Ultimately, correctly identifying emerging environmental degradation patterns is imperative to maintaining ecological balance and protecting public health in urban communities. By training models on available datasets of regional water and air quality indicators, a Pearson correlation coefficient calculator provides a data-driven approach to not only accurately identify the areas with environmental deterioration but also pinpoint the specific features - suspended particulate matter (SPM) and fecal coliform - that most directly lead to pollution and other problems. 

In conclusion, the hypothesis was proven true as a comprehensive machine learning pipeline was developed that could assess overall environmental quality by combining air and water quality data, while achieving high accuracy. The pipeline involved key steps of preprocessing the raw data, aligning and concatenating the datasets, labeling each data point with a categorical quality rating, training an ensemble model, and performing feature selection. Through this process, a robust classifier was created that demonstrated strong performance in distinguishing between favorable and challenged environments. The feature correlations uncovered provide actionable insights into the primary factors tied to environmental degradation. By fusing disparate datasets and leveraging advanced machine learning, this work enables data-driven monitoring and diagnosis of ecological health. The pipeline and models constructed lay the groundwork for developing predictive tools to guide policies and interventions to safeguard communities. Incorporating additional data sources and applying techniques like deep learning could further enhance predictive accuracy. Overall, this research highlights the potential of AI to drive evidence-based decisions for environmental sustainability.
\subsection{Future Works}
The current predictive modeling pipeline for environmental quality provides a robust starting point. However, to enhance model accuracy and comprehensiveness, incorporating diverse complementary metrics capturing additional ecological perspectives is necessary. Specifically, adding agricultural data such as soil health, crop patterns, and vegetation would provide insights into a critical human impact area. Climate and weather data could help control for periodic and geographic variations. Biodiversity statistics lend a biological view of ecosystem stability. Furthermore, integrating social and economic factors like demographics and development would offer a human lens regarding the societal drivers and implications of environmental challenges. Blending these multidimensional datasets within the flexible modeling framework constructed here would allow for a more nuanced, holistic understanding of intricate ecological issues, leading to more targeted, effective sustainability policies and interventions. This future work highlights the power of an interdisciplinary data science approach that unifies statistical and domain expertise to address society’s most pressing environmental concerns \cite{Guidotti2017}.

Another area for future improvement is enhancing the machine learning techniques used. While the current pipeline relies on conventional methods like random forests and support vector machines, adopting modern deep-learning approaches could improve performance. Graph neural networks could precisely better model the complex interdependencies in ecological systems by representing relationships between entities like species, nutrients, and pollutants. Recurrent networks could identify temporal dynamics by analyzing time series data. Multi-modal models fusing numeric and image data could incorporate satellite imagery to assess vegetation and land use changes. Additionally, active learning techniques could be leveraged to achieve more accurate labeling and reduce the labeled data needed. By selectively sampling uncertain and informative points to be manually labeled, far fewer labeled examples would be required. Furthermore, transfer learning from related tasks could provide proper initialization and representations to enable training with limited environment-quality data. In summary, applying cutting-edge profound learning advances in architecture, temporal modeling, multi-modality, active learning, and transfer learning could significantly boost model capabilities.

Another critical area for future work is model interpretation and result visualization. While predictive accuracy is essential, generating trusted and actionable insights from AI is equally crucial. Techniques like LIME could highlight the features most influential to predictions, enabling an understanding of model behavior. Interactive visualizations could be developed to allow filtering and drilling down on regions and factors of concern. Automated report generation summarizing project conclusions using natural language techniques would make results more accessible to diverse stakeholders. Critically evaluating results for potential biases and uncertainties is essential to ensure fair, transparent conclusions. Overall, substantial opportunities exist to enhance model interpretability, interactive analysis, and responsible, transparent reporting to convert predictions into meaningful, trustworthy insights for sustainability decision-making.

One exciting area for future development is applying the insights from this modeling work to design effective filtration systems that can actively remove key contaminants degrading environmental quality. Since influential factors like nitrate, phosphate, and particulate matter have been identified through feature selection, we can strategically target filtration mechanisms for these harmful substances. For example, the model results suggest that deploying nitrate and phosphate adsorbent materials at agricultural runoff sites could have an outsized impact on water quality. Additionally, high-efficiency particulate air (HEPA) filtration systems in industrial facilities could help lower ambient particulate matter levels in surrounding communities. By quantitatively linking contaminants to environmental quality deterioration, we can develop technical remediation solutions tailored to address the most impactful pollution sources. New filtration materials and technologies like nanofibers and reactive membranes show promise for removing specific threats identified by the model. An interdisciplinary approach combining cutting-edge data science with expertise in environmental engineering can help design the next generation of intelligent, sustainable filtration infrastructure to improve ecological health actively.

Another important next step is to integrate this predictive modeling pipeline with real-time sensing and monitoring infrastructure. The current model provides powerful batch analytics on static historical data. However, dynamic real-time environment quality assessment could be enabled by connecting these models to live sensor feeds. Sensors for parameters identified as highly predictive by the model, like nitrate levels or particulate matter, could be deployed in bodies of water or air quality monitoring stations. The incoming sensor data could then be fed into the optimized AI models to generate frequent environment quality updates, alerts for deteriorating conditions, and anticipate future states.

Furthermore, connecting the pipeline outputs to visualization dashboards would allow stakeholders to monitor overall environmental health continually and dive into granular details as needed. Investing in this end-to-end infrastructure would transform broad batch analytics into precisely-targeted, responsive decision support. Policymakers could rapidly respond to emerging threats before they escalate. Scientists could discern the environmental impacts of perturbations in near real-time. Ultimately, transitioning the advanced analytics developed here into an operationalized system would maximize its value for continuous sustainability progress.

\bibliographystyle{IEEEtran}
\end{document}